\pdfoutput=1

\documentclass[11pt]{article}

\usepackage[]{eacl2023}

\usepackage{times}
\usepackage{latexsym}

\usepackage[T1]{fontenc}

\usepackage[utf8]{inputenc}

\usepackage{microtype}

\usepackage{amsmath}
\usepackage{amssymb}
\usepackage{xspace}
\usepackage{algorithm}
\usepackage[noend]{algpseudocode}[1]
\usepackage[normalem]{ulem}
\usepackage{multicol}
\usepackage{booktabs}
\usepackage{adjustbox}
\usepackage{comment}
\usepackage{pifont}
\usepackage{array,graphicx}

\definecolor{cadmiumgreen}{rgb}{0.0, 0.42, 0.24}

\newcommand{\triple}[3]{\texttt{<#1, #2, #3>}}
\newcommand{\smatch}[0]{\textsc{Smatch}\xspace}

\newcommand{\smatchpp}[0]{\textsc{Smatch++}\xspace}

\newcommand{\hclimb}[0]{\xspace\includegraphics[trim={1cm 1cm 0 0cm}, scale=0.08]{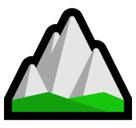}}

\newcommand*\rot{\rotatebox{90}}

\title{\smatchpp: Standardized and Extended Evaluation of Semantic Graphs}

\author{Juri Opitz \\
Heidelberg University \\
  \texttt{opitz.sci@gmail.com} 
}
\begin{document}
\maketitle
\begin{abstract}
The \smatch metric is a popular method for evaluating graph distances, as is necessary, for instance, to assess the performance of semantic graph parsing systems. However, we observe some issues in the metric that jeopardize meaningful evaluation. E.g., opaque pre-processing choices can affect results, and current graph-alignment solvers do not provide us with upper-bounds. Without upper-bounds, however, fair evaluation is not guaranteed. Furthermore, adaptions of \smatch for extended tasks (e.g., fine-grained semantic similarity) are spread out, and lack a unifying framework. 

For better inspection, we divide the metric into three modules: pre-processing, alignment, and scoring. Examining each module, we specify its goals and diagnose potential issues, for which we discuss and test mitigation strategies. For pre-processing, we show how to fully conform to annotation guidelines that allow structurally deviating but valid graphs. For safer and enhanced alignment, we show the feasibility of optimal alignment in a standard evaluation setup, and develop a lossless graph compression method that shrinks the search space and significantly increases efficiency. For improved scoring, we propose standardized and extended metric calculation of fine-grained sub-graph meaning aspects. Our code is available at \url{https://github.com/flipz357/smatchpp}

\end{abstract}

\section{Introduction}

Semantic graphs such as meaning representations (MRs) aim at capturing the meaning of a text. Typically, these graphs are rooted, directed, acyclic, and labeled. Vertices denote semantic entities, and edges represent semantic relations (e.g., \textit{instrument}, \textit{cause}, etc.). A prominent MR framework is \textit{Abstract Meaning Representation} \textit{(AMR}), proposed by \citet{banarescu-etal-2013-abstract}, which anchors in a propositional knowledge base \cite{palmer-etal-2005-proposition}. 

Using a metric such as \smatch \cite{cai-knight-2013-smatch}, we can measure a distance (or similarity) between graphs, by aligning nodes, and counting matching graph triples. In fact, \smatch measurement has various applications. It is used for selecting parsing systems that project AMR structures \cite{flanigan-etal-2014-discriminative, may2017semeval, xu-etal-2020-improving, hoang2021ensembling, bevilacqua2021one} and various other semantic graphs \cite{van-noord-etal-2018-evaluating, zhang-etal-2018-cross, oepen2020mrp, stengel-eskin-etal-2020-universal, martinez-lorenzo-etal-2022-fully, lin-etal-2022-neural}, for MR-based evaluation and diagnostics of text generation systems \cite{opitz-frank-2021-towards, manning-schneider-2021-referenceless, ribeiro-etal-2021-structural, hoyle-etal-2021-promoting}, as backbone in an ensemble parsing algorithm \cite{NEURIPS2021_479b4864ensemble}, and for studying cross-lingual phenomena \cite{uhrig-etal-2021-translate, wein-etal-2022-effect}. Through  \smatch measured on sub-graphs, we can assess similarity of linguistic phenomena such as semantic roles, negation, or coreference \cite{damonte-etal-2017-incremental}, a property that can be leveraged in neural text embeddings \cite{opitz-frank-2022-sbert}.
 
However, \smatch measurement is non-trivial and lacks specification. For instance, \smatch involves an NP-hard \textit{optimization problem} of structural \textit{graph alignment}, which distinguishes it from most metrics used in other evaluation tasks. In practice, a solution of this problem is found by employing a hill-climber. However, a hill-climber terminates at local optima, and it cannot inform us about a score upper-bound. In the end, this means that we lack information about the quality of the returned solution, potentially lowering our trust in the final evaluation. To mitigate this issue, we would like to study the possibility of optimal solution, or solution with a tight upper-bound. There are also other issues, on which we lack understanding. E.g., we do not know to what extent different pre-processing choices may affect the evaluation results, and we miss specification of \smatch's popular fine grained sub-graph metrics \cite{damonte-etal-2017-incremental}, where it is unclear how sub-graphs should be best extracted and compared.

\paragraph{Paper structure and contributions} First, we describe and generalize the \smatch metric (\S \ref{sec:gensmatch}), and summarize recent \smatch variants in one framework. Then we break the metric down into three modules (\S \ref{sec:pipeline}), which ĺets us better distribute our attention over its key components. For each module, we discuss specification of goals and mitigation of issues. In the pre-processing module  (\S \ref{sec:pp1}), we motivate graph standardization to allow safer matching of equivalent MR graphs with different structural choices. In the optimization module (\S \ref{sec:pp2}), we test strategies for solving the alignment problem with optimality guarantees. In the scoring module (\S \ref{sec:pp3}), we discuss standardized and extended scoring of fine-grained semantic aspects, such as causality, tense, and location.

\section{Related work}

\paragraph{Metric standardization} An inspiration for us is the work of \citet{post-2018-call}, who propose the popular \textsc{SacreBleu} framework for fairer comparison of machine translation systems with a standardized \textsc{Bleu} metric  \cite{papineni-etal-2002-bleu}. Specifically, \textsc{SacreBleu} ships \textsc{Bleu} \textit{together with} a specified tokenizer -- prior to this, BLEU differences between systems could depend on different tokenization protocols. Facing the challenging problem of graph evaluation, a main contribution of our work is that we i) analyze weak spots in the current evaluation setup and ii) discuss ways of mitigating these issues, aiming at best evaluation practices.

\paragraph{MR metrics} \citet{cai-lam-2019-core} introduce a variant of \smatch \cite{cai-knight-2013-smatch} that penalizes dissimilar structures if they are situated in proximity of the graph root, motivated by their assumption that `core-semantics' are located near the root of MR graphs. Furthermore, \citet{10.1162/tacl_a_00329} introduce a \smatch variant that performs a graded match of semantic concepts (e.g., \textit{cat} vs.\ \textit{kitten}), aiming at extended use-cases beyond parsing evaluation, where MRs of different sentences need to be compared. Similarly, \citet{wein-schneider-2022-accounting} adapt an embedding-based variant of \smatch for cross-lingual MR comparison. We show that the different \smatch adaptions can be viewed through the same lens with a generalized notion of triple match. Furthermore, \citet{damonte-etal-2017-incremental} propose fine-grained \smatch that measure MR agreement in different aspects, such as \textit{semantic roles}, \textit{coreference} or \textit{polarity}. We diagnose and mitigate issues in the aspectual assessment, and show how to extend the measured aspects.

Conceptually different MR metrics have been proposed by \citet{anchieta2019sema} and \citet{song-gildea-2019-sembleu} who aim at increased efficiency using structure extraction via breadth-first traversals, or \citet{opitz-etal-2021-weisfeiler} who compare MRs of different sentences with Wasserstein Weisfeiler-Leman kernels \cite{weisfeiler1968reduction, NEURIPS2019_73fed7fd}. Since significant parts of this paper are independent from \smatch-specific scoring\footnote{E.g., input standardization (\S \ref{sec:pp1}) and sub-graph extraction for fine-grained aspectual matching (\S \ref{subsec:fine-grained}).}, other MR metrics can profit from our work.

\section{\smatch: Overview and generalization}
\label{sec:gensmatch}

We introduce \smatch  and define a generalized \smatch, so that we can summarize recent \smatch variants in one framework.

\paragraph{Preliminary I: MR graph} If not mentioned otherwise, we view an MR graph $a$ as a set of triples, where a triple has one of two types. Unary triples have the structure \triple{x}{:rel}{c}, where the source \texttt{x} is a variable and the target \texttt{c} is a descriptive label that shows the type or an attribute of $x$, depending on the edge label \texttt{:rel}.\footnote{E.g., \triple{x}{:instance}{cat} would indicate that `\texttt{x} is a cat', while \triple{y}{:polarity}{-} means $y$ is negated.} Using variables such as \texttt{x} we can (co-)refer to different events and entities and capture complex events. Binary triples have the structure \triple{x}{:rel}{y}, where both the source \texttt{x} and the target \texttt{y} are variables.\footnote{E.g., \triple{x}{:location}{y}, which means that $x$ is located at $y$, or \triple{x}{:arg0}{y} which usually indicates that $y$ participates as the agent in the event referred to by $x$.}

\paragraph{Preliminary II: \smatch} The idea of \smatch is to measure structural similarity of graphs via the amount of triples that are shared by $a$ and $b$. To obtain a meaningful score, we must know an alignment $map$:$~vars(a) \leftrightarrow vars(b)$ that tells us how to map a variable in the first MR to a variable in the second MR. In this alignment, every variable from $a$ can have at maximum one partner in $b$ (and vice versa). Let an application of a $map$ to a graph $a$ be denoted as $a^{map}:=\{t^{map}~;~ t \in a\}$, where  $t^{map}$ of a triple $t = \triple{x}{:rel}{y}$ is set to  $t^{map} = \triple{map(x)}{:rel}{map(y)}$ for binary triples, and $t^{map} = \triple{map(x)}{:rel}{c}$ for unary triples. 

Under any alignment $map$, we can calculate an overlap score $f$. In original \smatch, $f$ is the size of the triple overlap of $a$ and $b$:
\begin{equation}
\label{eq:hardscore}
f(a, b, map) = |a^{map} \cap b|.
\end{equation},

Ultimately we are interested in  

\begin{equation}
\label{eq:smatch}
     F = \max_{map} f(a, b, map),
\end{equation}

Finding a maximizer $map^\star$  lies at the heart of \smatch, and we will dedicate ourselves to it later in \S \ref{sec:pp2}. For now, we assume that we have $map^\star$ at our disposal. Therefore, we can calculate \textit{precision} ($P$) and \textit{recall} ($R$):

\begin{align}
\label{eq:pr}
    P = |a|^{-1} F,~~~~~~~R = |b|^{-1} F,
\end{align}

to obtain a final F1 evaluation score: $2PR/(P+R)$. With such a score, we can assess the similarity of MRs, and compare and select parsing systems. 

\paragraph{Generalizing \smatch} In \smatch, two triples are said to match if they are identical under a mapping. I.e., we match with $match(t, t') := I[t = t']$ that returns 1 if two triples $t$ and $t'$ are the same, and zero else (we omit the $map$ for simplicity). Recently, \smatch has been adapted and tailored to different use-cases. E.g., \smatch has been extended to incorporate word embeddings \cite{10.1162/tacl_a_00329, wein-schneider-2022-accounting} to match \triple{x}{:instance}{c} triples for studying cross-lingual MRs or MRs of different sentences.\footnote{Consider \triple{x}{:instance}{cat} extracted from one sentence vs.\ \triple{y}{:instance}{kitten} extracted from another sentence. A graded match is required to properly assess the similarity of the concepts.} On the other hand, \citet{cai-lam-2019-core} propose a root-distance bias, based on the assumption that `core-semantics' lie in the proximity of an MR's root. 

We find that we can summarize such variants in one framework. We achieve this by introducing \textit{a scaled triple matching} function: 

  $$
match(t, t')= w_t^{t'}\cdot 
\begin{cases}
I[t=t'],  \text{~~~~~ if } t_2, t'_2 \neq \text{:inst.}\\
I[t_1=t'_1] \cdot sim(t_3, t_3')  \text{~~ else}
\end{cases}
$$

For matching concepts with embeddings, we can use an embedding similarity on the descriptive concept labels with $sim(c, c')$ and the importance weight $w_{t}^{t'} = 1~ \forall t,t'$.\footnote{That is, if the triples are not instance triples, we check whether the triples are equivalent (as in standard \smatch), but if both triples are instance relation triples and the variables $t_1, t'_1$ are set to equal each other, we calculate the similarity between their descriptive concept labels.} For Root-distance biased \smatch as proposed by \citet{cai-lam-2019-core} we set $w_t^{t'}$ such that we discount triple matches that are distant to the root.\footnote{For a properly normalized final score if $\exists~(t,t'), w^t_t\neq 1$, we may have to change denominators in Eq.\ \ref{eq:pr}}

Our generalization does not change or constrain the original \smatch. Instead, our goal was to define a more general framework of \smatch-type metrics that unifies recently proposed \smatch variants and show possibilities for further extension. For the following studies, we set \smatchpp to basic \smatch, which is recovered by setting ~$\forall t, t': w_{t}^{t'} = 1$ and $sim(c,c') := I[c = c']$.

\section{A modular view on \smatch}
\label{sec:pipeline}

To set the stage for inspection, we break \smatch down into three modules. i) \textit{Preprocessing}, ii) \textit{Alignment}, and iii) \textit{Scoring}. In particular, i) \textit{Preprocessing} discusses any graph reading and processing in advance of the alignment. ii) \textit{Alignment} revolves around the search mechanism used for finding an optimal mapping $map^\star$. iii) \textit{Scoring} involves calculating final scores and statistics that are returned to a user. For each module, we will specify its goals, assess potential weak spots and discuss mitigation.

\section{Module I: Pre-processing}
\label{sec:pp1}

\subsection{Module goal and current implementation}
 
\begin{figure}
    \centering
    \includegraphics[width=0.9\linewidth]{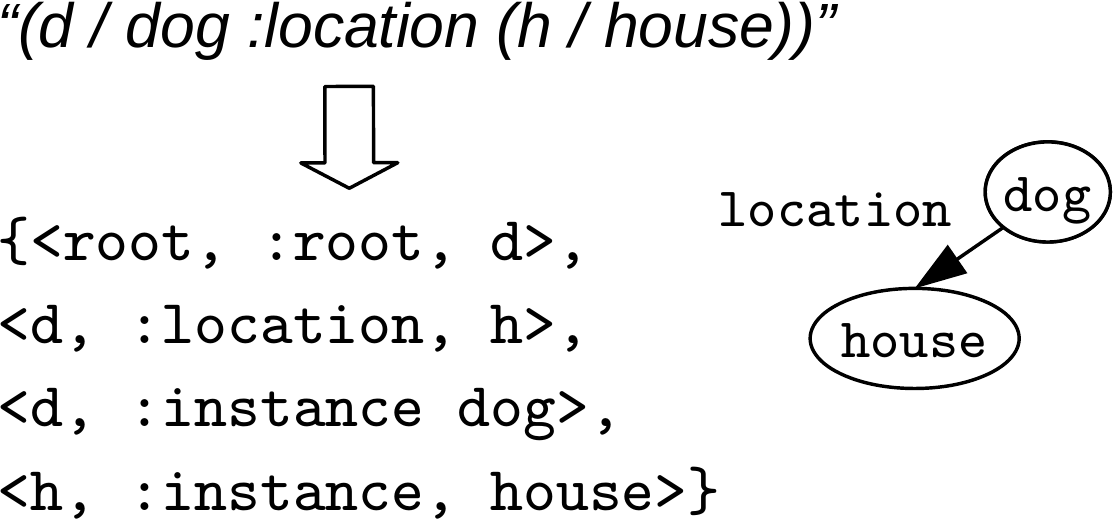}
    \caption{A serialized MR string is read into a graph.}
    \label{fig:read}
\end{figure}

MRs are typically stored and distributed in a `Penman' string format, which can serialize any rooted and directed graph into a string. The goal of this module is to project two serialized textual MRs onto two sets of triples, as outlined in Figure \ref{fig:read}.

The target domain of this projection should be a \textit{standardized} MR graph space, where format divergences that do not impact graph semantics are eliminated. Original \smatch performs pre-processing as follows: i) lower-case strings, ii) de-invert edges (e.g., \triple{x}{:relation-of}{y} $\rightarrow$ \triple{y}{:relation}{x}). However, while these steps seem sensible, more steps can be undertaken to enhance evaluation. 

\subsection{Two structures, one meaning: reification}
\label{subsec:reify}
\begin{figure}
    \centering
    \includegraphics[width=0.9\linewidth]{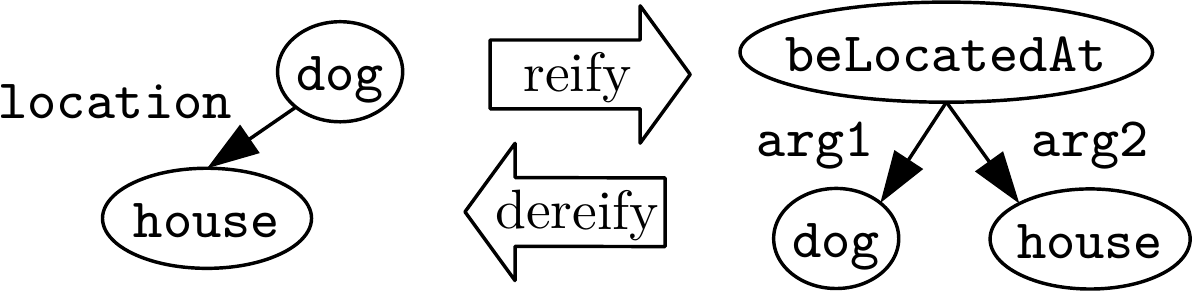}
    \caption{Outline of \textit{location}-reification.}
    \label{fig:rfy}
\end{figure}

Some MR guidelines, including the AMR guideline, allow meaning-preserving structural graph translations \cite{guide2013, goodman1} with so called \textit{reifications} (or \textit{de-reification} as an inverse mechansim). A subset of relations is selected to constitute a semantic relation core set (e.g., \texttt{:arg0}, \texttt{:arg1}, ..., \texttt{:op1}, \texttt{:op2},...) and for all other remaining relations (e.g., \texttt{:location}, \texttt{:time}), we use rules to map the relation to a sub-graph, where the rule-triggering relation label is projected onto a node, and the former source and target of the relation are attached with outgoing core relations. E.g., consider Figure \ref{fig:rfy}, where a reification is applied to a \triple{x}{:location}{y} relation. In this case, the rule is:

\begin{itemize}
    \item location (de)reififcation: \\ \triple{x}{:location}{y} \\ $\iff$ \\ \triple{z}{:instance}{beLocatedAt} \\ $\land$ \triple{z}{:arg1}{x} \\ $\land$ \triple{z}{:arg2}{y},
\end{itemize}

 where \texttt{:arg1} indicates the thing that is found at a location \texttt{:arg2}. 
 
 The question whether an annotator should use either means of representation, is answered in the guidelines as follows: \textit{whenever they feel like it} \cite{guide2013}. Therefore, a parser should not be penalized or rewarded for projecting reified (or non-reified) structures. 

\paragraph{Empirical assessment of effect} To understand the effect that reification can have on the final \smatch score, it is interesting to study an edge-case: evaluating graphs that are fully reified against graphs that are fully de-reified. As a data set we take LDC2017T10, a standard AMR benchmark. Additionally, we gather automatic parses by applying an AMR parser \cite{xu-etal-2020-improving}. 

The results of this experiment are shown in Table \ref{tab:reifyvsdereify}. In the first three lines (i-iii) we compare \textit{equivalent} translated versions of the test partition (gold vs.\ gold). We find that two equivalent gold standards can be judged to be very different (73.9 points, -26.1 points). A similar phenomenon can be observed when looking at the parses. The best parser score is achieved when comparing parses and references in the domain of reified graphs (82.8 points). On the other hand, if only the reference is reified, the parser score drops by 20 points (viii). 

However, we also see that the results of a basic evaluation (vii) is practically the same as the result when evaluating with de-reified graphs (vi), indicating that both parser and gold annotation abstain from reification, where possible.

\begin{table}
    \centering
    \scalebox{0.83}{
    \begin{tabular}{|l|ll|rr|}
   \toprule
    &\multicolumn{2}{|c}{Data setup}&\multicolumn{2}{|c|}{\smatch} \\
    &X & Y&  Orig & rfyStd \\
     \toprule
        i) & gold dereify & gold reify & 73.8 &100.0 \\
        ii) & gold standard& gold reify & 73.9 &100.0 \\
        iii) & gold standard& gold dereify & 100.0 &100.0 \\
        \midrule
        iv) & parser dereify& gold reify & 60.9 & 82.8\\
        v) & parser reify& gold reify & 82.8 &82.8\\
        vi) & parser dereify& gold dereify & 81.4 & 82.8\\
        vii) & parser standard& gold standard & 81.4 &82.8\\
        viii) & parser standard& gold reify & 60.9 &82.8\\
        ix) & parser standard& gold dereify & 81.4 &82.8\\
        \bottomrule
    \end{tabular}}
    \caption{Results of meaning-preserving translations. rfyStd: score when we project X and Y into standardized reified space.}
    \label{tab:reifyvsdereify}
\end{table}

\paragraph{Discussion} Having established that rule-based graph translations can enhance evaluation fairness, we pose the question: \textit{should we prefer reification or de-reification for space standardization}? 

The answer should be \textit{reification}, since it can be seen as a form of generalization. More precisely, we note that reification of non-core relations is \textit{always} possible. In fact, an interesting effect of reified structures is that they equip us with the means to attach further structure, or features, to semantic relations. On the other hand, however, de-reification is not always possible. It is only well-defined if there is no incoming edge into the node that corresponds to the non-core relation\footnote{It is not clear to which node the incoming edge (that now does not have a target) should be re-attached: the \texttt{arg0} or \texttt{arg1} of the outgoing edges of the former node? Either choice would likely come with a change in meaning.}, and if there are not more than two outgoing edges\footnote{I.e., since reification can potentially be used to model $n$-ary relations, only in the case where $n=2$ we can model the structure with a single (labelled) edge}. 

However, there are also (practical) arguments against reification. Consider that de/non-reified MRs are smaller and have more edge label differentiation. This i) may facilitate more intuitive display for humans and ii) shrinks the alignment search space. Indeed, a large solution space may have ramifications for evaluation optimality and efficiency (in \S \ref{sec:pp2}, we empirically study this issue). Therefore, when taking into account that the empirical effect size appears neglectable in the average case, these trade-offs may not always be justified, and we may instead use de-reification, where possible.

\subsection{Triple removals}

Duplicate triples are triples that occur more than once. We find that they are sometimes produced by some parsers. Additionally, some parsers introduce a node more than once, which results in two triples \triple{x}{:instance}{a} and \triple{x}{:instance}{b}. Currently, \smatch removes all such introductions of a second concept, but does not remove duplicate triples. By contrast, we propose to remove all duplicate triples, since they have no clear semantics, and stay agnostic to second introductions of a concept (in some MRs, it may be acceptable that an entity is the instance of two concepts), keeping all such triples (if they are not identical).\footnote{Due to rare occurrence of such phenomena in our parsed data, we find the effects of either choice to be negligible.}

\section{Module II: Alignment}
\label{sec:pp2}

The goal of this module is solving Eq.\ \ref{eq:smatch}, finding a $map^\star$ for optimal matching score. 

\smatch uses a hill-climber for solving Eq.\ \ref{eq:smatch}. An issue with this is that such a heuristic terminates at local optima and cannot provide us with any \textit{upper-bounds}. Upper-bounds, however, can inform users about the \textit{quality} of the outputted solution and thus increase the trustworthiness of the final score (and any parser comparison that is based thereupon). Therefore, we can conclude that using a hill-climber seems \textbf{practical but may not be optimal}, especially when considering cases where fair comparison needs to be \textit{guaranteed}. Instead, we would like to use an Integer Linear Program (ILP) to obtain the (optimal) solution. Alternatively, at least, we would like to know a tight upper-bound to inform ourselves about the trustworthiness of the final score. But ILP is NP hard, and therefore it seems \textbf{optimal but possibly not practical}, a conception that might favor the usage of a hill-climber. 

Triggered by these considerations, we review the hill-climber and the ILP and assess their effects on MR evaluation, with two desiderata in mind: evaluation quality and efficiency. Additionally, we propose a strategy for loss-less MR compression that can improve efficiency of any solver.

\begin{figure}
    \centering
    \includegraphics[width=\linewidth]{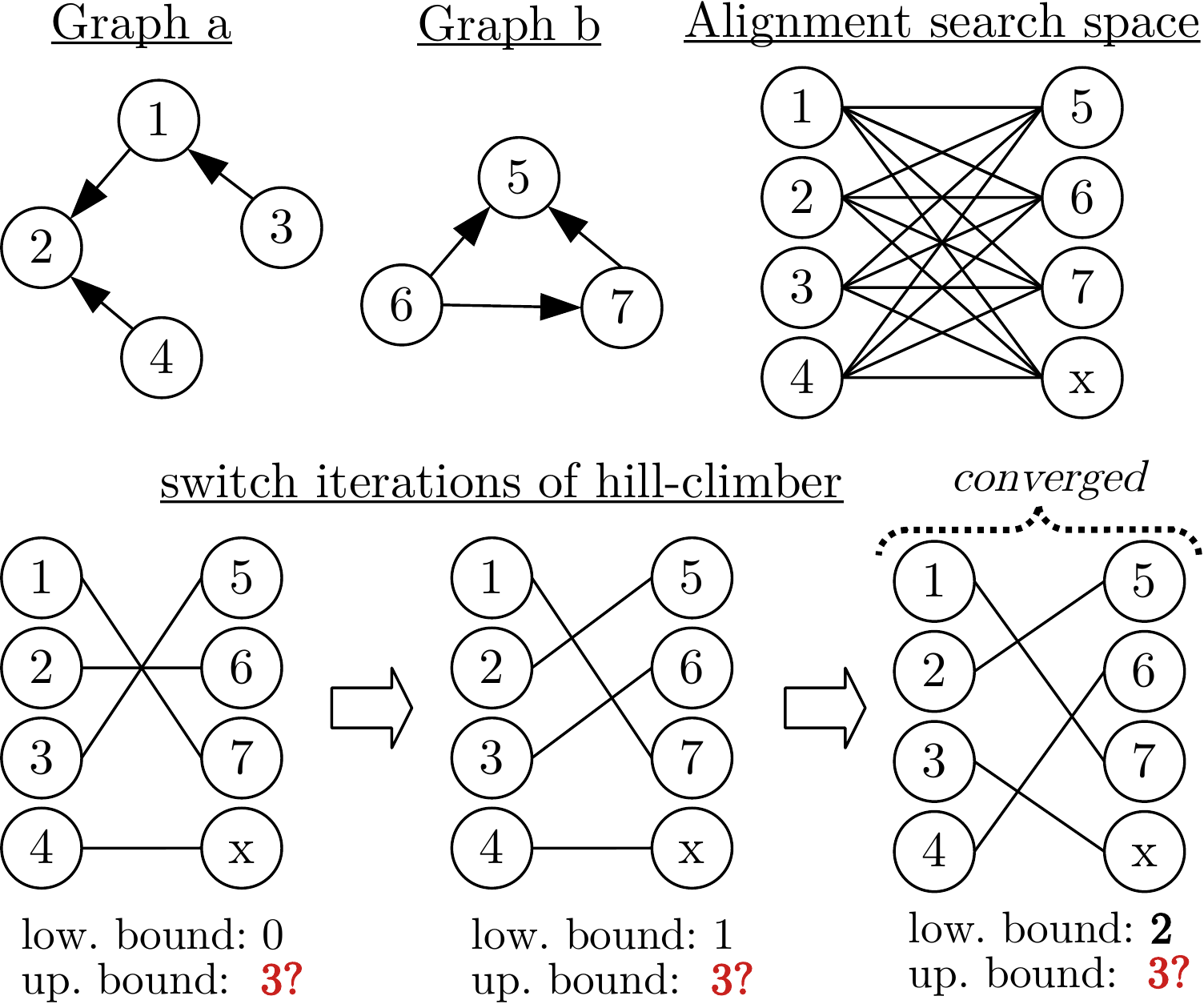}
    \caption{Sketch of search space (top) and hill-climber run (bottom). Every hill-climber step constitutes an improved lower bound, but we cannot obtain a tight upper-bound (an accessible trivial upper-bound is the amount of triples in the smaller of two graphs: 3).}
    \label{fig:align}
\end{figure}

\subsection{Practical but not optimal: hill-climber} 

\smatch hill-climbing uses two operations, which we denote as \textit{switch}, and \textit{assign}. The \textit{assign}-operation assigns a variable from $vars(a)$ to an unaligned variable from $vars(b)$: $(i, \emptyset) \rightarrow (i, j=map'(i))$, where $map'$ is a candidate map. The switch operation does an alignment cross-over with respect to two alignment pairs, i.e.: $(i, j=map(i)) \land (k, l=map(k)) \rightarrow (i, l=map'(i)) \land (k, j=map'(k))$, where $map$ is the current alignment and $map'$ the candidate alignment. In each iteration, we examine all possible \textit{switch}- and \textit{assign} options, and greedily choose the best one.\footnote{\textit{Assign} is just a special instance of the more general \textit{switch} so we can ablate the \textit{assign} step. Then, \textit{assign} becomes  $(i, \emptyset=map(i)) \land (k, j=map(k)) \rightarrow (i, j=map'(i)) \land (k, \emptyset=map'(k))$, which is a \textit{switch}.}   An example alignment procedure is shown in Figure \ref{fig:align}.

In practice, we can resort to multiple random restarts, to find better optima. However, this hardly addresses the underlying issue: we lack any information on upper-bounds, which may decrease trustworthiness of results, especially when facing larger graphs with lots of local optima.

\subsection{ILP: Optimal, but less practical?}
\label{subsec:compress}

We would like to use Integer Linear Programming (ILP) for optimal solution of the graph alignment. 

\paragraph{Problem statement} Assume two graphs $g, g'$ with node sets $V, V'$. Let $u(i, j)$ denote the amount of unary triple matches, given we align $i$ from $V$ to $j$ from $V'$, counting matches of triples that involve one MR variable. On the other hand, $b(i, j, k, l)$ will denote the amount of structural binary triple matches, given we align $i$ from $V$ to $j$ from $V'$  and $k$ from $V$ to $l$ from $V'$.  Here, we count matching binary triples that involve two MR variables. Usually, these data are pre-computed. Let $x$ indicate our current $map$, i.e., if $x_{ij} = 1$ then we align $i$ from $V$ to $j$ from $V'$. We find our solution at

\begin{equation*}
\begin{array}{ll@{}ll}
\text{max}  & \displaystyle\sum\limits_{\substack{(i, j)  \in \\ V \times V}} u(i, j) x_{ij} + \displaystyle\sum\limits_{\substack{(i, j, k, l) \in\\ (V \times V)^2}} b(i,j,k,l)x_{ij}x_{kl}\\
\text{st} & \displaystyle\sum\limits_{j} x_{ij} \leq 1;~~~\displaystyle\sum\limits_{i} x_{ij} \leq 1\\
 & x_{ij} \in \{0,1\} ~~~ \forall (i, j) \in V \times V'
\end{array}
\end{equation*}

The constraint ensures that every node from one graph is aligned, at maximum, to one node from the other graph. By linearization, and introducing structural variables $y$, we obtain the equivalent ILP:

\begin{equation*}
\begin{array}{ll@{}ll}
\text{max}  & \displaystyle\sum\limits_{\substack{(i, j)  \in \\ V \times V'}} u(i, j) x_{ij} + \displaystyle\sum\limits_{\substack{(i, j, k, l) \in\\ (V \times V')^2}} b(i,j,k,l)y_{ijkl}\\
\text{st} & \displaystyle\sum\limits_{j} x_{ij} \leq 1;~~~\displaystyle\sum\limits_{i} x_{ij} \leq 1\\
& y_{ijkl} \leq x_{ij},~~~~~ \forall (i, j, k, l) \in (V \times V')^2\\
& y_{ijkl} \leq x_{kl},~~~~~ \forall (i, j, k, l) \in (V \times V')^2\\
 & x_{ij} \in \{0,1\} ~~~~~ \forall (i, j) \in V \times V' \\
 & y_{ijkl} \in \{0,1\} ~~~ \forall (i, j, k, l) \in (V \times V')^2,
\end{array}
\end{equation*}

where the structural variables, if active, show us countable binary triple matches. This is an NP complete problem, imposing limits on its capability to provide us with optimal solutions for larger graphs (note, however, that we can retrieve intermediate solutions and upper-bounds).  

\subsection{Reduced search space with lossless graph compression} 

We observe that in an MR $a$, every variable $x \in vars(a)$ is related to a concept $c$, e.g., \triple{x}{:instance}{cat}. This means that a concept $c$ does \textit{identify} a variable $x \in vars(a)$ iff $\forall y \in vars(a):\triple{y}{:instance}{c} \Rightarrow y = x$. Therefore, if $x$ denotes a $cat$, and there is no other entity in the MR that also denotes a cat, then $x$ may be referred to simply by $cat$. This carries over to pairs of MRs: which are the focus of the paper -- instead of considering $vars(a)$, we simply consider $vars(a) \cup vars(b)$. Therefore, we can replace all $n$ variables from $vars(a) \cup vars(b)$ that are \textit{identified} by concepts, with the corresponding concepts (see Appendix \ref{app:reduce} for a full example). This shrinks the search space by reducing the amount of variables that the optimizer has to consider. Note that such a compression is \textit{lossless}, in the sense that the possibility of full reconstruction of the original MR is ensured. This implies that if two compressed MRs are assessed as (non-)isomorphic, then the uncompressed MRs are also (non-)isomorphic.

\subsection{Solver experiments}

Two questions are of main interest: 1. \textit{RQ1, solution quality}: (How) do the final \smatch results depend on the solver? 2. \textit{RQ2, solution efficiency}: How does the evaluation time depend on the solver? In addition, we would like to assess how our answers to RQ1 and RQ2 might be affected by reification (resulting in a bigger search space) and MR compression (resulting in a smaller search space). 

\paragraph{Setup} We simulate a standard AMR parsing evaluation setting. We parse the LDC2017T10 testing data with six parsers: $\mathcal{P}_1$ \cite{xu-etal-2020-improving}, $\mathcal{P}_2$ \cite{cai-lam-2020-amr}, $\mathcal{P}_3$ \cite{lindemann-etal-2020-fast}, $\mathcal{P}_4$ \cite{zhang-etal-2019-amr}, $\mathcal{P}_5$ \cite{lyu-titov-2018-amr}, $\mathcal{P}_6$ \cite{cai-lam-2019-core}. We evaluate the parsers using ILP or hill-climber (denoted by \hclimb). As is standard, we show F1 micro corpus scores. For reference, we also run evaluation with the standard \smatch hill-climbing script (denoted as \textit{previous}). We observe that we successfully reproduce the scores from the standard \smatch script with our \hclimb~ implementation (first two lines of Table \ref{tab:res}).\footnote{An improvement is obtained for $\mathcal{P}_5$. We find that we can mostly attribute this to a bug in the original script that prevents proper graph reading of some parses of $\mathcal{P}_5$.}

\subsubsection{RQ1: solution quality}\label{subsec:ilpsolverresults}

\begin{table*}
    \centering
    \scalebox{0.83}{
    \begin{tabular}{lrrrrrrrr|rll}
       & & &\multicolumn{6}{c}{parser scores (ranked)} & time & \# vars & quality\\
     & data & optim &$\mathcal{P}_1$ &$\mathcal{P}_2$ &$\mathcal{P}_3$ &$\mathcal{P}_4$ &$\mathcal{P}_5$ & $\mathcal{P}_6$ & secs & (tot., avg., max.) & (yield, bound)\\
     \midrule
        & basic & prev.\ & 81.4$_{(1)}$ & 80.3$_{(2)}$& 77.0$_{(3)}$ & 76.3$_{(4)}$ & 74.5$_{(5)}$ & 73.1$_{(6)}$ & 50.4 & (20346, 15, 129) & (217702, +?)\\
       &  basic & \hclimb$_4$ & 81.4$_{(1)}$  & 80.3$_{(2)}$& 77.0$_{(3)}$& 76.2$_{(4)}$& 75.1$_{(5)}$ & 73.1$_{(6)}$ & 49.9 & see above & (217716, +?)\\
      &  basic & ILP & 81.5$_{(1)}$  & 80.4$_{(2)}$& 77.1$_{(3)}$& 76.5$_{(4)}$& 75.2$_{(5)}$ &  73.3$_{(6)}$ & 98.0 & see above & (218072, +0)\\
        \cmidrule{2-12}
       \rot{\rlap{all vars}} &  reify & \hclimb$_4$  &82.8$_{(1)}$ &81.3$_{(2)}$&78.3$_{(3)}$&77.7$_{(4)}$&76.9$_{(5)}$ &74.7$_{(6)}$&134.5 & (27812, 20, 174) & (288597, +?)\\
       &  reify & ILP &  83.5$_{(1)}$  & 82.1$_{(2)}$ & 79.3$_{(3)}$ & 78.7$_{(4)}$ & 77.7$_{(5)}$ & 75.8$_{(6)}$ & 300.5 & see above & (291370, +13)\\
        \midrule
        \midrule
       &   basic & \hclimb$_1$  &72.9$_{(1)}$  & 70.9$_{(2)}$  & 67.1$_{(3)}$  & 66.2$_{(4)}$  & 64.6$_{(5)}$  &  61.5$_{(6)}$  & 7.3& (5568, 4, 62) & (74163, +?)\\
      &   basic & ILP & 73.3$_{(1)}$  & 71.3$_{(2)}$ & 67.5$_{(3)}$ & 66.3$_{(4)}$ & 65.0$_{(5)}$ & 62.1$_{(6)}$ & 11.7 & see above & (75036, +0) \\
        \cmidrule{2-12}
       &  reify & \hclimb$_1$  & 74.9$_{(1)}$  & 72.8$_{(2)}$  & 69.5$_{(3)}$  & 68.7$_{(4)}$  & 67.7$_{(5)}$  & 64.6$_{(6)}$  & 31.4 &  (10704, 8, 106) & (124323, +?)\\
       \rot{\rlap{~~compress}}&  reify & ILP & 76.7$_{(1)}$  & 74.1$_{(2)}$ & 71.3$_{(3)}$ & 70.6$_{(4)}$ & 69.5$_{(5)}$ & 66.4$_{(6)}$ &27.3 & see above & (129019, +0)\\
        
        \bottomrule
    \end{tabular}}
    \caption{Parser evaluation. \textit{time} refers to the approximate total time needed to evaluate a single parser (i.e., processing 1371 graph pairs). ~~\includegraphics[trim={1cm 1cm 0 0cm},scale=0.08
        ]{pics/10292oigbr8y5.png}-$N$ indicates hill-climber optimizer with $N$ restarts. \textit{quality}: solution quality of solver -- first number is the amount of matching triples summed over all six parser evaluations (yield); second number indicates the tightest found upper-bound (which is only known by ILP).}
    \label{tab:res}
\end{table*}

\paragraph{Insight: Better alignment $\rightarrow$ safer evaluation} Importantly, we see that the ILP yields score increments for all parsers, which signals the occurrence of alignment problems, where the \hclimb (despite multiple restarts) did not find the optimal solution. The effect-size is larger for reified graphs. We find differences of up to 1 point F1 score (Table \ref{tab:res}: reify \hclimb$_4$  vs.\ reify ILP). This can be explained by the growth of the alignment search space -- reification makes graphs larger and introduces more MR variables. This explanation is further supported by contrasting the amount of unique final objective values against the size of the alignment space with different random initializations of the hill-climber (Appendix \ref{app:hc20inits}, Figure \ref{fig:solquality}). We see that i) for many graph pairs there are multiple local optima, and ii) the likelihood of finding a non-global optimum with the \hclimb~ increases for larger/reified graphs.

We further study upper-bounds and solution quality (right column of Table \ref{tab:res}). The ILP found the optimal solution in all cases, yielding 218072 matching triples. The \hclimb$_4$ finds 217,700 matching triples (99.83\%), which misses the mark by 350 triples. When evaluating reified graphs, the ILP returns 291370 matches and thus misses its temporary tightest upper-bound by 13 triples, indicating that in a few cases, a sub-optimal solution might have been found.\footnote{Indeed, we find one graph by $\mathcal{P}_2$, and one graph by $\mathcal{P}_6$, where the ILP terminates after a 240s timeout that we set, and returns a temporary solution.} The \hclimb$_4$, however, yields only 288,597 matches (99.04\%) and misses the temporary ILP upper-bound by 2,786. The growing gap underlines the degrading quality of the hill-climber when facing larger graphs. 

Finally, the (slight) \textit{differences} in increments among parsers when we evaluate them on reified graphs indicate that different parsers do make different decisions on when to reify an edge. For instance the score difference $\Delta$ for reified graphs vs.\ non-reified graphs (using ILP) of $\mathcal{P}_5, \mathcal{P}_6$ is 2.5 points, for $\mathcal{P}_1$ 2 points and for $\mathcal{P}_2$ 1.7 points. This supports our theoretical insights from \S \ref{subsec:reify} -- reification can make parser comparison fairer.

\subsubsection{RQ2: Solution efficiency}

\paragraph{Insight I: ILP isn't that impractical} It seems to be commonly presumed that original \smatch uses a hill-climber to make evaluation more practical and fast. However, our results qualify this presumption.  For evaluating a full corpus (1371 graph pairs), \smatch with ILP needs only about 48 seconds longer than original \smatch with hill-climber (50s vs 98s). When the search space grows (due to reification) the time gap widens to a difference of 165 seconds. However, the consistent improvement of scores due to ILP (signaling sub-optimal hill-climber solutions) can make the time increase acceptable for evaluations where fairness is critical.

\paragraph{Insight II: MR compression increases evaluation speed} Viewing the last four rows of Table \ref{tab:res}, we see that the MR compression i) did not lead to switched system ranks and ii) increased the evaluation speed by a large factor. Using MR compression, the ILP runs a full system evaluation in 11.7 seconds for standard graphs and 27.3 seconds for the reified graphs. Given that the MR compression is lossless (c.f.\ \S \ref{subsec:compress}), it provides us with an option for more efficient evaluation that is also safe (i.e., optimal).

\section{Module III: scoring}
\label{sec:pp3}

\subsection{Main scores: Precision, Recall and F1}

\begin{table}
    \centering
    \scalebox{0.7}{
    \begin{tabular}{llllllll}
    \toprule
      &\multicolumn{6}{c}{parser scores} \\
     avg.\ & $\mathcal{P}_1$ &$\mathcal{P}_2$ &$\mathcal{P}_3$ &$\mathcal{P}_4$ &$\mathcal{P}_5$ & $\mathcal{P}_6$   \\
     \midrule
         mic.\  & 81.5$_{80.7}^{82.2}$& 80.4$_{79.6}^{81.2}$ &  77.1$^{77.8}_{76.2}$ & 76.5$_{75.6}^{77.2}$& 75.2$_{74.5}^{75.8}$ & 73.3$^{74.1}_{72.4}$\\
         mac.\ & 82.6$^{83.3}_{81.8}$ &81.4$_{80.7}^{82.1}$ & 79.0$^{79.8}_{78.2}$  & 78.3$_{77.5}^{79.1}$ & 76.2$^{77.0}_{75.4}$  & 75.9$^{76.6}_{75.0}$\\
        \midrule
    \end{tabular}}
    \caption{Evaluation with additional macro statistics and confidence intervals. Solver: ILP.}
    \label{tab:macro-confidence}
\end{table}

The goal of this module is to provide the user with a final result. As discussed in \S \ref{sec:gensmatch}, the main scores (Precision, Recall, and F1) follow directly from the $map^\star$. The final score is typically micro averaged, summing matching statistics across all graph pairs before they are normalized. \smatchpp makes two additions, macro-scoring and confidence intervals. Macro-averaging scores over graph pairs can be a useful complementary signal, specifically when comparing high-performance parsers \cite{opitz-frank-2022-better}. Additionally, we adopt the bootstrap assumption \cite{efron1992bootstrap} for calculating confidence intervals. To make calculation feasible, bootstrapping is performed after the alignment stage. Table \ref{tab:macro-confidence} shows results of the additional statistics. Confidence intervals range between +-[0.5, 1] points for all parsers. Macro score shows an outlier, where $\mathcal{P}_6$ (+2.6 points) is more positively affected than other parsers (+[1.0, 1.9] points).\footnote{We find a potential explanation in a motivation of $\mathcal{P}_6$'s creators to focus on semantics in proximity of an MR's top node (the proportion of such semantics increases when the graph is smaller, and smaller graphs have more influence on macro average than on micro average).}

\subsection{Measuring aspectual semantic similarity}

We observe considerable interest in applying fine-grained aspectual MR metrics \cite{damonte-etal-2017-incremental} for inspecting linguistic aspects captured by MRs (e.g., semantic roles, negation, etc.). Applications range from parser diagnostics \cite{lyu-titov-2018-amr, xu-etal-2020-improving,  bevilacqua2021one, martinez-lorenzo-etal-2022-fully}, to NLG system diagnostics and sentence similarity \cite{opitz-frank-2021-towards, opitz-frank-2022-sbert}. Formally, given an aspect of interest $asp$ and an MR $g$, we apply a subgraph-extraction function $sg(g, asp)$ to build an aspect-focused sub-graph, and compute a matching score (e.g., F1).

\paragraph{Review of previous implementation} We study the description in \citet{damonte-etal-2017-incremental} and the most frequently used implementation \cite{smimpl}. The treated aspects\footnote{See Appendix \ref{app:aspects} for a full overview.} are divided in two broad groups: \textbf{i) alignment-based matching}: For some aspects, we extract aspect-related genuine sub-graphs, on which we calculate an optimal alignment. \textbf{ii) bag-of-label matching:} for other aspects, we detect aspect-related variables and gather associated node labels\footnote{I.e., from \triple{x}{:instance}{label} triples} in a bag/list, to compute an overlap score based on simple set intersection. 

E.g., \textit{SRL}-aspect belongs to the first category (\textbf{i}): we extract \triple{x}{:arg$_{n}$}{y} relations, and their corresponding \textit{instance} triples (here: \triple{x}{:instance}{c}, and \triple{y}{:instance}{c'}). Then we calculate \smatch on such SRL-subgraphs. The \textit{Negation}, \textit{Named Entity} (NEs) and \textit{Frames} aspect is put into the second group (\textbf{ii}). We look for a relation/node-label that signals a particular aspect, e.g., \triple{x}{:polarity}{-} (for negation) or \triple{x}{:name}{y} (for NEs), we extract \texttt{x}, and replace \texttt{x} with the descriptive label \texttt{c} from \triple{x}{:instance}{c}.  For \textit{Frames}, we search for \triple{x}{:instance}{c} where \texttt{c} is a PropBank predicate, and collect \texttt{c}. Finally, we can evaluate without an alignment, using set intersection.

\paragraph{Open questions} We pose two questions:

\begin{enumerate}
    \item Can the sub-graph extraction be improved?
    \item Are there other aspects that we can measure?
\end{enumerate}

\subsection{Improving sub-graph extraction}
\label{subsec:fine-grained}

\begin{figure}
    \centering
    \includegraphics[width=\linewidth]{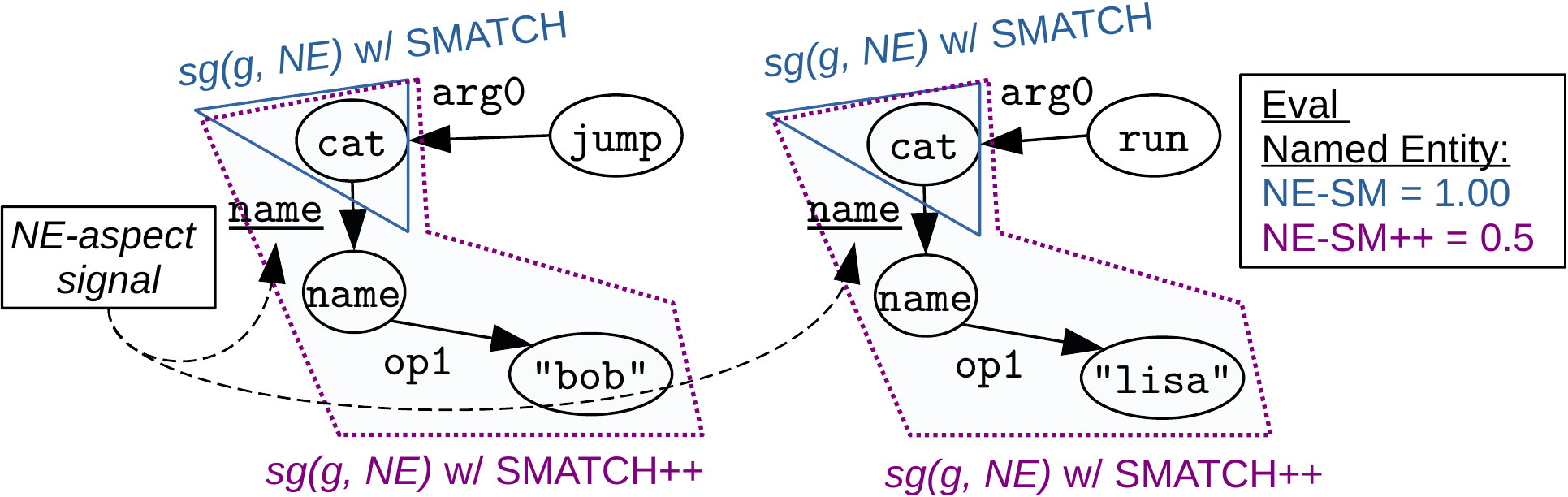}
    \caption{Named Entitiy (NE) sub-graph extraction with \smatch vs.\ \smatchpp}
    \label{fig:ne_aspect}
\end{figure}

\paragraph{Sensible range of extraction} For some phenomena, the current extraction range is clearly too limited. For instance, let us consider named entities, which can be captured in more complex and nested MR structure. E.g., in AMR, one node typically indicates the type of the named entity (NE), and another multi-node structure represents its name and other attributes. Consider two AMRs $a$ and $b$, from which we want to extract NE structures to measure the agreement of the graphs w.r.t.\ NE similarity. As shown in Figure \ref{fig:ne_aspect}, assume that one graph is about \textit{a cat named Bob}\footnote{Triples: \triple{x}{:instance}{cat}, \triple{y}{:instance}{name}, \triple{x}{:name}{y}, \triple{y}{:op1}{"bob"}.}, while the other graph is about \textit{a cat named Lisa}\footnote{Triples: \triple{x}{:instance}{cat}, \triple{y}{:instance}{name}, \triple{x}{:name}{y}, \triple{y}{:op1}{"lisa"}.}. Obviously, the MRs have similarities in their NE structure (since there are named cats), but also differences (since the cats have different names). However, NE-focused \smatch only extracts \textit{cat} and \textit{cat}, and returns maximum score. 

Hence, for all finer-grained aspects that are captured by non-atomic MR structures (e.g., Named Entities), we propose to gather the full sub-graph starting at the aspect-indicating relation or node label. In the NE example, as shown in Figure \ref{fig:ne_aspect}, we would be provided a score of 0.5, better reflecting the similarity of the two NE structures. 

\paragraph{Sub-graph compression, align and match} We find a middle-ground in the advantages of the coarse matching (concreteness, efficiency) and graph alignment (safe matching) by using alignment with lossless MR compression. This is optimal and efficient, and alleviates the need to switch among fine and coarse extraction methods.

\subsection{Extending fine-grained scores}

\begin{figure}
    \centering
    \includegraphics[width=\linewidth]{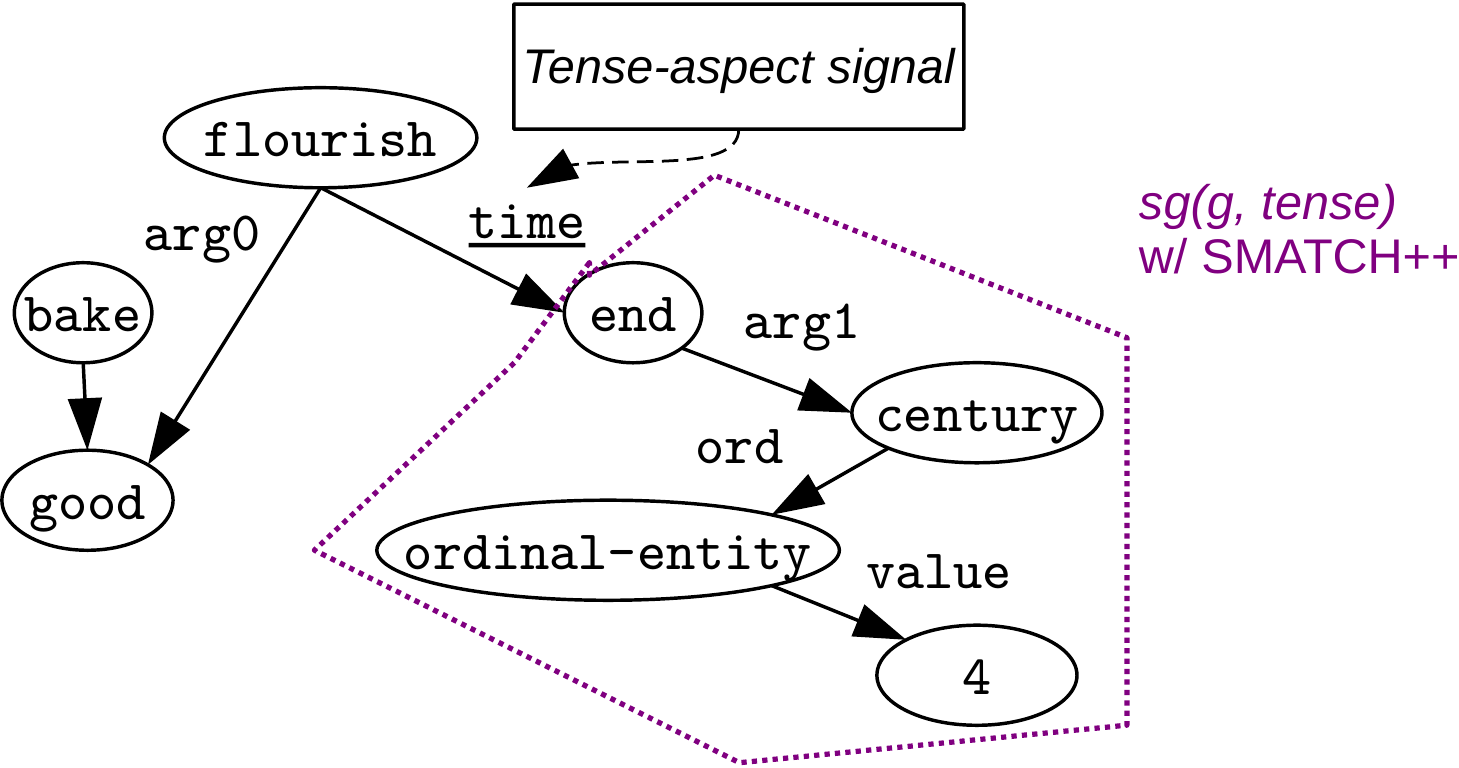}
    \caption{Temporal sub-graph extraction with \smatchpp for an MR capturing \textit{``Baked goods flourished at the end of the fourth century''}.}
    \label{fig:tense_extract}
\end{figure}

 \paragraph{Beyond negation and named entities -- other semantic aspects} We find that the fine-grained \smatch metrics by \citet{damonte-etal-2017-incremental} miss some interesting features captured by MRs. For instance, four interesting AMR aspects that are currently not captured are \textit{cause}, \textit{location}, \textit{quantification}, and \textit{tense}. \smatchpp allows their integration in a straightforward way. An example for tense extraction is displayed in Figure \ref{fig:tense_extract}, where our \smatchpp sub-graph extraction extracts the complete temporal sub-graph, triggered by the edge label \texttt{:time} (if we would resort to the style of fine-grained \smatch, we would miss larger parts of the temporal structure, only extracting the node label \textit{end}).

Results of fine-grained parser diagnostics for \textit{cause}, \textit{location}, \textit{quantification}, and \textit{tense} are shown in Table \ref{tab:subres}.

Interestingly, we see that projecting \textit{causality} seems hard: all parsers tend to struggle when assessing causal structures (31.2 up to 47.8 F1 points), showing much room for improvement. The temporal structures, on the other hand, can be assessed with somewhat higher accuracy (48.4 up to 67.7 points). We also see  some switched ranks, indicating different parser strengths. Overall, parser score differences seem notably more pronounced than when calculating \smatch(++) on the full graphs, showing the difficulty of capturing finer phenomena, and highlighting strengths of more recent parsers.

\begin{table}
    \centering
    \scalebox{0.83}{
    \begin{tabular}{llllllll}
    \toprule
      &\multicolumn{6}{c}{parser scores} \\
     aspect & $\mathcal{P}_1$ &$\mathcal{P}_2$ &$\mathcal{P}_3$ &$\mathcal{P}_4$ &$\mathcal{P}_5$ & $\mathcal{P}_6$   \\
     \midrule
         cause  & 47.8& 47.4&  44.4 & 35.7& 31.4 & 31.2\\
         location & 61.8 &53.2$\downarrow$ & 54.7$\uparrow$  & 49.2$\downarrow$ & 51.7$\uparrow$  & 40.0\\
         quant  & 69.4 & 67.4 & 58.4 & 56.8 & 56.5 & 55.8 \\
         tense   &  67.7 & 62.3 & 58.5 & 56.5 & 50.3 & 48.4\\
        \midrule
    \end{tabular}}
    \caption{Evaluation for \textit{causal} and \textit{temporal} structures. $\downarrow\uparrow$ indicate switched ranks. Solver: ILP.}
    \label{tab:subres}
\end{table}

\section{Conclusion}

\smatchpp is the first specification of a standardized, extended, and extensible \smatch metric. We aim at i) standardized and transparent comparison of graph parsing systems, and ii) improved extensibility for custom applications.\footnote{See Appendix \ref{app:bestpractice} for a summary of the default setup.} The applications can include finer parser diagnostics and measuring semantic sub-graph similarities such as \textit{quantification}, \textit{cause}, or \textit{tense} with our fine-grained metrics.

\section*{Acknowledgments}

We thank our reviewers for their helpful feedback.

\section*{Limitations}

We have to leave some questions open. First, we would have liked to shed more light on the solvers' behaviors when facing large graphs, in isolation. On one hand, our benchmark corpus indeed contains some large MRs with many variables, including reified MRs and MRs that represent multiple sentences (up to 174 variables, cf.\ Table \ref{tab:res}). We have shown that ILP could cope with these harder problems, providing optimal solutions in reasonable time. When facing bigger graphs, however, we can expect that the solution quality of the hill-climber quickly degrades, while the ILP will struggle to find optimal solutions. While our graph compression strategy can help mitigate this issue by reducing the alignment search space, it would be interesting to study the quality of temporary solutions, or of solutions of LP relaxation. There are also relaxed ILP solvers \cite{klau2009new} that iteratively tighten the lower and the upper-bound. They could prove useful for aligning larger MR graphs, or, at least, to find useful upper-bounds.

Second, in this paper we studied \smatch(++) that measures \textit{structural overlap} and assigns each triple the \textit{same weight}. But structural differences of similar degree can have a different impact on overall meaning similarity as perceived by humans, which can have ramifications for measuring sentence similarity \cite{opitz-etal-2021-weisfeiler} and meaningful evaluation of strong AMR parsers \cite{opitz-frank-2022-better}. Therefore, for a deeper assessment of MR similarity we may have to use conceptually different metrics, or explore \smatchpp-based strategies and (sensibly) weigh triples depending on label importance or compose an overall score by weighting measured sub-aspect similarities.

\bibliography{anthology, custom}
\bibliographystyle{acl_natbib}

\appendix

\section{Appendix}
\label{sec:appendix}

\subsection{Lossless graph pair reduction example}
\label{app:reduce}

Consider \textit{the cat scratches another cat}: $a$=$\big\{\triple{s}{:instance}{scratch},$ 
$\triple{c}{:instance}{cat},$
$\triple{d}{:instance}{cat},$
$ \triple{s}{:arg0}{c},$ 
$\triple{s}{:arg1}{d}\big\}$ \\ and \textit{the gray cat scratches the small plant}:\\
$b$=$\big\{\triple{x}{:instance}{scratch},$ 
$\triple{y}{:instance}{cat},$
$\triple{z}{:instance}{plant},$
$\triple{w}{:instance}{small},$
$\triple{v}{:instance}{gray},$
$\triple{x}{:arg0}{y},$ 
$\triple{x}{:arg1}{z},$
$\triple{y}{:mod}{v},$ 
$\triple{z}{:mod}{z}\big\}$.

The lossless compression is $a'$=$\big\{\triple{c}{:instance}{cat},$
$\triple{d}{:instance}{cat},$
$ \triple{scratch}{:arg0}{c},$ 
$\triple{scratch}{:arg1}{d}\big\}$ and $b'$=$\big\{\triple{y}{:instance}{cat},$
$\triple{scratch}{:arg0}{y},$ 
$\triple{scratch}{:arg1}{plant},$
$\triple{y}{:mod}{gray},$ 
$\triple{plant}{:mod}{small}\big\}$.

The alignment search space is reduced from a size of more than 100 candidates to 2 candidate options ($y=c$, or $y=d$).

\subsection{Assessing solution quality variability in dependence of variables}\label{app:hc20inits}

\begin{figure}
    \centering
    \includegraphics[width=\linewidth]{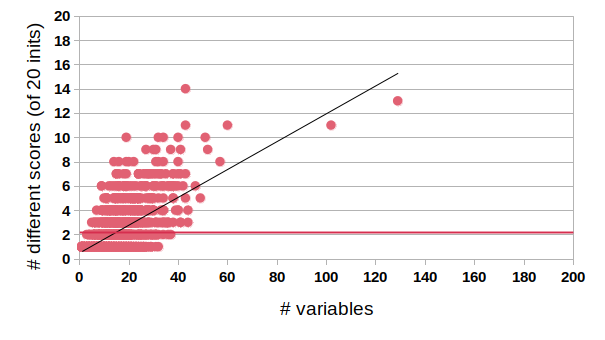}
    \includegraphics[width=\linewidth]{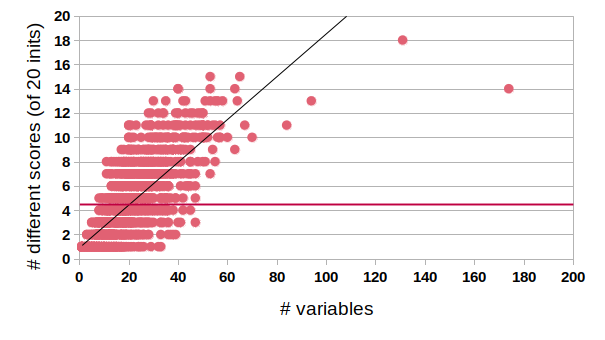}
    \caption{Assessing solution quality variability. Top: basic graphs, bottom: reified graphs. Diagonal line: linear trend. Horizontal line: arithmetic mean. See text in \S \ref{app:hc20inits} for more description and \S \ref{subsec:ilpsolverresults} for discussion.}
    \label{fig:solquality}
\end{figure}

We use the parses of an example parser ($\mathcal{P}_5$)\footnote{We ran the experiment also with parses from other systems but always ended up with essentially the same results}. For every evaluation pair, we re-start the hillclimber 20 times, and collect the scores related to the found local optima. The Y-axis in Figure \ref{fig:solquality} shows the amount of unique scores found among the 20 tries (note that there could be more unique alignments that would result in the same score -- this is not captured in this Figure). The X-axis shows the amount of alignment variables. In different terms, a higher point in this Figure is equivalent to a larger pool of local optima of different quality, and thus we can conjecture a greater likelihood that the optimal solution is not returned by the hill-climber.

\subsection{Aspect overview}
\label{app:aspects}

\paragraph{Previously measured aspects} 

For all aspects we retrieve F1, Precision, and Recall.
\begin{enumerate}
\item Measured under alignment
\begin{enumerate}
    \item SRL: extract \triple{x}{arg$_n$}{y} triples and corresponding instance triples.
    \item Coreference/Re-entrancies: extract \triple{x}{rel}{y} triples for which there is another triple \triple{z}{:rel'}{y} (meaning y is a re-entrant node) and also extract corresponding instance triples.
\end{enumerate}
\item Measured via bag-of-structure extraction and set operations
\begin{enumerate}
    \item Concepts: collect all node labels.
    \item Frames: collect all node labels where the label is a PropBank predicate frame.
    \item NonSenseFrames: see above, but with sense label removed
    \item NE: Named entities, collect all node labels that have an outgoing \texttt{:name} relation.
    \item Negation: collect all node labels that have an outgoing \texttt{:polarity} relation.
    \item Wikification: collect all node labels that have an incoming \texttt{:wiki} relation.
    \item IgnoreVars: replace all variables in triples with concepts, collect triples.
\end{enumerate}

\end{enumerate}
SRL, Named Entities, coreference (re-entrant nodes)

Additional aspects measured by us: \textit{Cause}, \textit{Tense}, \textit{Location}, \textit{Quantifier}.

\paragraph{We change:} Add default option for extracting aspect sub-graphs, measure all aspects under alignment.

\paragraph{Aspects we added:} 
\begin{itemize}
    \item \textit{Cause}: Cause is modeled via \texttt{cause-01}. We extract label of \texttt{:arg1} (what is caused?) and subgraph of \texttt{:arg2}, the cause itself.
    \item \textit{tense}: Tense is modeled via \triple{x}{:time}{y} edge. We extract label of \texttt{the thing that happens} and subgraph of \texttt{y}, the temporal description where it happens.
 \item \textit{location}: Similar to above but with \texttt{:location} edge.
\item \textit{quantifier}: Similar to above but with \texttt{:quant} edge.
    
\end{itemize}

\subsection{Best practice}
\label{app:bestpractice}

To provide a balance between efficiency, safety and meaningfulness of scores, default procedure of \smatchpp is currently set to:

\begin{enumerate}
    \item Pre-processing: lower-casing, duplicate-removal, de-reify where applicable.
    \item Alignment: Solver: ILP. Triple-match: $w_t^t=1~\forall t,t'$;~~ $sim(c, c'):=I[c=c']$ 
    \item Scoring: Precision, Recall, F1, Bootstrap confidence intervals 
\end{enumerate}

An option to increase efficiency without incurring a loss in safety and meaningfulness is achieved by adding graph compression to the pre-processing. It is set as the default for fine semantic aspect scores. Also, to ensure utmost safety, we have to consider applying reification standardization (incurring a significantly longer evaluation time).

\end{document}